\documentclass[runningheads]{llncs}
\usepackage{graphicx}
\usepackage{amsmath,amssymb} 
\usepackage{color}
\usepackage[width=122mm,left=12mm,paperwidth=146mm,height=193mm,top=12mm,paperheight=217mm]{geometry}
\begin{document}
\pagestyle{headings}
\mainmatter
\def\ECCV18SubNumber{1689}  

\title{MGGAN: Solving Mode Collapse using \\ Manifold Guided Training} 



\author{Duhyeon Bang \and Hyunjung Shim}
\institute{School of Integrated Technology, Yonsei University, South Korea}

\maketitle

\begin{abstract}
Mode collapse is a critical problem in training generative adversarial networks. To alleviate mode collapse, several recent studies introduce new objective functions, network architectures or alternative training schemes.  However, their achievement is often the result of sacrificing the image quality. In this paper, we propose a new algorithm, namely a manifold guided generative adversarial network (MGGAN), which leverages a guidance network on existing GAN architecture to induce generator learning all modes of data distribution. Based on extensive evaluations, we show that our algorithm resolves mode collapse without losing image quality. In particular, we demonstrate that our algorithm is easily extendable to various existing GANs. Experimental analysis justifies that the proposed algorithm is an effective and efficient tool for training GANs. 
\end{abstract}


\section{Introduction}

Generative adversarial networks (GANs) \cite{ref20} are a family of generative models that implicitly estimate the data distribution in an unsupervised manner. This is accomplished by learning to generate new data samples instead of explicitly constructing a density function. Because GANs do not rely on strong statistical assumptions or constraints on distributions, there are no performance limitation on modeling complex manifolds of data distribution. Owing to this attractive nature, GANs have been successful in image generation tasks; various studies report that their image quality is superior to the traditional generative models in that GANs produce sharp and realistic images.

Despite promising achievements, GANs are notoriously hard to train due to the training instability and sensitivity to hyperparameters. Training instability causes two problems: poor image quality and lack of image diversity. These two issues are in a trade-off relationship with each other. Existing studies aim to improve either image quality or diversity. In this paper, our primary interest is to improve image diversity without sacrificing image quality.

The lack of image diversity in GAN training is also known as mode collapse, in which ${\mathit{P}}_{model}$ captures a single or few major modes of ${\mathit{P}}_{data}$ while ignoring many small modes. To address this problem, we propose a novel algorithm, namely the manifold guided generative adversarial network (MGGAN), which integrates a newly proposed guidance network to the existing GAN architecture. Note that the standard GAN consists of a discriminator network and generator network. The discriminator aims to distinguish the fake images produced by the generator from real images. Meanwhile the generator aims to fool the discriminator by generating fake images as realistic as possible. On this standard GAN architecture, we leverage the guidance network, which encourages generator to learn all modes of ${\mathit{P}}_{data}$. The goal of the guidance network is to teach generator in that ${\mathit{P}}_{model}$ matches ${\mathit{P}}_{data}$ in the learned manifold space. For that, the guidance network consists of an encoder for manifold mapping and a discriminator for measuring the dissimilarity between the distributions of ${\mathit{P}}_{data}$ and ${\mathit{P}}_{model}$ in the manifold space. 
In this way, the characteristics of learned manifold space is reflected in generator training. To solve mode collapse, we employ an encoder layer of a pre-trained autoencoder to define manifold mapping. This autoencoder is optimized for reconstructing all samples of real images, and fixed after pre-training so that it is not updated during GAN learning. Because this autoencoder learns to represent all training data \cite{ref34}, the manifold learned by the encoder is effective to represent all modes of ${\mathit{P}}_{data}$.

The contributions of our MGGAN is summarized as follows.
\begin{enumerate}
    \item The manifold space derived by the autoencoder represents the most critical features for reconstructing all modes of ${\mathit{P}}_{data}$. The guidance network provides feedback to generator in a way that the distribution of critical features is well restored in the fake images. In this way, we induce the generator to learn all modes of ${\mathit{P}}_{data}$, thereby producing diverse samples.
    \item Because our encoder network is pre-trained and fixed during GAN training, it prevents the errors of encoder training from propagating to the generator and discriminator training.
    \item There is no range or unit mismatch between the loss of discriminator and that of guidance network because both networks use adversarial loss. As a result, training of our network is stable. 
    \item The proposed algorithm resolves mode collapse without sacrificing the image quality. 
\end{enumerate}


\section{Background}\label{sec02}
 A variety of techniques have been proposed in the past for solving mode collapse, and can be categorized into following two groups.

\subsection{Regularizing the discriminator for resolving the mode collapse}
Although the standard GAN \cite{ref20} theoretically proves that generative modeling can be formulated by minimizing the Jensen-Shannon divergence ($JSD$), authors recommend the non-saturated GAN for the actual implementation \cite{ref20,ref08}. The non-saturated GAN is designed to minimize $KL({\mathit{P}}_{model}\mathit{||}{\mathit{P}}_{data}) -2JSD$ for generator update, which holds a property of the reverse Kullback-Leibler (KL) divergence between ${\mathit{P}}_{data}$ and ${\mathit{P}}_{model}$ \cite{ref05}. Fedus et al. \cite{ref05} and Arjovsky et al. \cite{ref11} point out that the reverse KL-divergence is vulnerable to mode collapse in non-saturated GAN. Because the reverse KL evaluates dissimilarity between two distributions at every fake sample (i.e., ${\mathit{P}}_{model} (x) > 0$, for  all   $x$), there is no penalty for covering a fraction of the true data distribution. To address this issue, they suggest Wasserstein distance that holds the weakest convergence among existing GAN metrics. This new metric is effective in solving mode collapse by stabilizing GAN training. However, they approximate the Wasserstein distance by weight clipping, unfortunately causing a pathological behavior \cite{ref12}. On the contrary, D2GAN \cite{ref13} employs two antithetical discriminators: one minimizes forward KL divergence; the other minimizes reverse KL divergence. As a result, a generator produces samples to fool both discriminators simultaneously, and escape from mode collapse. However, they tend to increase instability because the goals of two antithetical discriminators conflict.

Unrolled GAN \cite{ref14} claims that mode collapse occurs because discriminator updates do not guarantees an optimal discriminator. Thus, they introduce a surrogate objective function that simulates a discriminator response to generator changes. Although their model is robust against mode collapse, it is not clear whether their achievement in mode collapse is the result of sacrificing the visual quality or not. Also, heavy computational complexity due to k-step discriminator updates is a well-known drawback of Unrolled GAN. 

DRAGAN \cite{ref15} states that non-convex loss function exhibits local minimax points, leading to mode collapse. Hence, authors propose a gradient penalty (GP) term in order to regularize sharp gradients. The GP term stabilizes GAN training, which is also effective in mitigating mode collapse. LSGAN \cite{ref28} replaces the sigmoid cross-entropy loss term used in standard GAN with a least squares loss term, which is equivalent to minimizing Pearson $\tilde{\chi}^2$ divergence. They assert that the replacement improves stability of learning process and reduces the possibility of mode collapse. However, existing studies, DRAGAN and LSGAN, do not provide a significant achievement for improving diverse image generation with real datasets. 

\subsection{Learning to map between the latent to data domain}

Mode collapse leads to ignoring minor modes of data distribution. To address this problem, several recent studies propose learning of a mapping function from ${\mathit{P}}_{data}$ to ${\mathit{P}}_z$; namely an inference mapping. ALI \cite{ref01} and BiGAN \cite{ref02} suggest a discriminator for joint distribution matching, which learns a relationship between data and latent distribution. This can be interpreted as ALI and BiGAN aim to recover a bidirectional mapping between the data and latent while standard GAN learns a unidirectional mapping, from ${\mathit{P}}_{z}$ to ${\mathit{P}}_{data}$; namely a generation mapping. However, their results do not improve existing GAN models. 

MDGAN \cite{ref03} and VEEGAN \cite{ref04} utilize a reconstruction loss as an additional constraint to the inference mapping. 
Although reconstruction loss is effective, its unit mismatches that of an adversarial loss; reconstruction loss is a distance measure while adversarial loss is a divergence measure. MDGAN separates training into a mode regularization step and a diffusion step, in order to reduce instability. 
Unlike MDGAN, VEEGAN applies the reconstruction loss in the latent domain rather than the data domain.
Hence, they mitigate the image quality degradation caused by the reconstruction loss applied in data domain (i.e. image blur). 

AGE \cite{ref16} suggests a new architecture composed of an encoder and a generator, and designs a adversarial learning between two networks without a discriminator. Because they do not rely on a discriminator, they greatly reduce the computational complexity and the converge time compared to the previous models utilizing a bidirectional mapping. 


\section{ Manifold guided generative adversarial networks}\label{sec03}

\begin{figure}[t!]
    \centering
    \includegraphics[height=4cm]{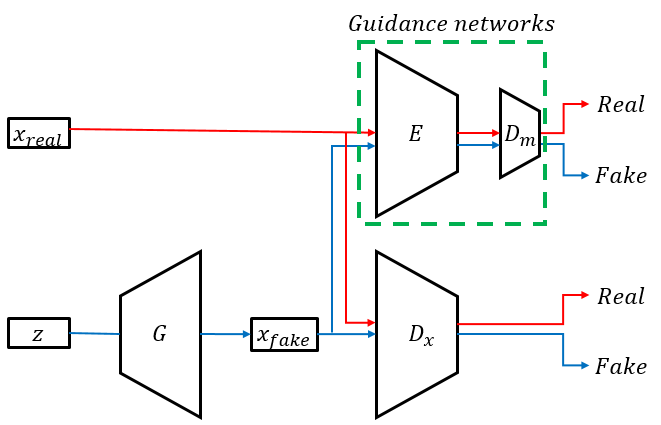}
    \caption{The proposed model Structure. $x_{real}$ and $x_{fake}$ are sample of ${\mathit{P}}_{data}$ and ${\mathit{P}}_{model}$, respectively; $z$ is latent vector; $E$, $G$, and  $D$ are the encoder, the generator, and the discriminator network.  The subscript of $D$ means input sample space. The guidance network consists of $e$ and $D_m$, where $m$ means manifold space. There are no difference with the standard GAN except adding our guidance network.}
    \label{fig01}
\end{figure}

Our goal is to generate diverse samples, i.e., solving mode collapse, without sacrificing the image quality. For that, we propose a new algorithm inducing a generator to learn all modes of ${\mathit{P}}_{data}$ as well as producing realistic samples.
Specifically, we introduce a guidance network, which guides the generator producing samples reflecting the manifold distributions. The standard GAN, which consist of a generator, $G$, and a discriminator, $D_{x}$, applied with this guidance network is shown in Fig.~\ref{fig01}.

For the sake of distinguishing between the true and the estimated probability distribution, we mark with a hat over estimated variables; in our study, since the encoder maps the true probability distribution to the manifold, $E(x\sim {\mathit{P}}_{data})$ is mapped onto ${\mathit{P}}_m$ and $E(x\sim {\mathit{P}}_{model})$ is mapped onto ${\mathit{P}}_{\widehat{m}}$, where $m$ represents the manifold space.

Our guidance network aims to reduce the divergence between the projection of ${\mathit{P}}_{data}$ and ${\mathit{P}}_{model}$ on the manifold space. The guidance network is composed of an encoder, $E$, and a discriminator, $D$. The encoder maps ${\mathit{P}}_{data}$ and ${\mathit{P}}_{model}$ to the manifold space. The discriminator for the guidance network, $D_m$, distinguishes the encoded ${\mathit{P}}_{model}$ from the encoded ${\mathit{P}}_{data}$, i.e., ${\mathit{P}}_{m}$ and ${\mathit{P}}_{\widehat{m}}$, respectively. The following equations show the objective function for our MGGAN, where the guidance network is implemented with the non-saturated GAN.

\begin{align}\label{eq01}
\begin{split}
    \mathop{\mathit{min}}_{D_x,D_m} \ \ 
    & \mathbb{E}_{x\sim {\mathit{P}}_{data}}\left[{\mathit{log} \left(D_x\left(x\right)\right) \ \ + \mathit{log}D_m\left(E\left(x\right)\right)\ }\right]  + \\&\mathbb{E}_{z\sim {\mathit{P}}_z} \left[\ {\mathit{log} \left(1-D_x\left(G\left(z\right)\right)\right)\ \ } + {\mathit{log} \left(1-D_m\left(E\left(G\left(z\right)\right)\right)\right)\ }\right]\ ,
\end{split}
\end{align}
\begin{equation}\label{eq02}
    {\mathop{\mathit{min}}_{G} -\ \mathbb{E}_{z\sim {\mathit{P}}_z}[{\mathit{log} \left(D_x\left(G\left(z\right)\right)\right)\ }+{\mathit{log} \left(D_m\left(E\left(G\left(z\right)\right)\right)\right)]\ . \ \ \ \ \ \ \ \ \ \ \ \ \ } }
\end{equation}

As described in Eq.\ref{eq01} and Eq.\ref{eq02}, two discriminators, $D_x$ and $D_m$, do not explicitly affect each other, but both of them influence the generator. From Eq.\ref{eq02}, the generator attempts to meet two goals simultaneously; the first is to minimize the dissimilarity between ${\mathit{P}}_{data}$ and ${\mathit{P}}_{model}$ equivalent to a non-saturated GAN, and the second is to minimize the dissimilarity between their mapped distributions onto manifold space. It is worthwhile noting that our two discriminators concurrently affect the generator training, and thus two discriminators are implicitly influenced each other through generator. Especially, the encoder of a guidance network is designed to derive the most representative manifold of $\mathit{P}_{data}$ where all modes of $\mathit{P}_{data}$ are captured.
As a result, the guidance network can induce the generator training in that the generator is capable of producing diverse samples, because ${\mathit{P}}_{model}$ encapsulates all modes of ${\mathit{P}}_{data}$ by reflecting the characteristics of the encoder of the guidance network.

\begin{figure}[t!]
    \centering
    \includegraphics[height=3.5cm]{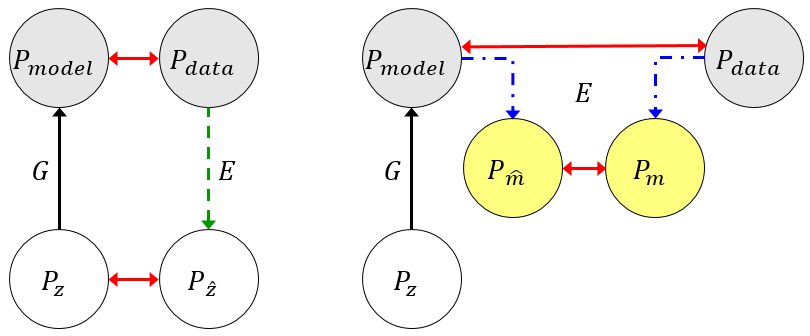}
    \caption{Difference between bidirectional mapping method and the proposed weakly bidirectional mapping algorithm. ${\mathit{P}}_{data}$ is data distribution. ${\mathit{P}}_z$ is simple prior distribution, generally uniform or Gaussian noise. ${\mathit{P}}_m$ is manifold distribution which we introduce on the proposed method. Hat means fake. The solid black line, the green dash line, and the blue dash dot line are generation, inference, and manifold mapping, respectively. $G$ and $E$ mean networks conducts each mapping. The double-sided arrow red line depicts divergence which GANs aims to reduce.}
    \label{fig02}
\end{figure}

\subsection{Compared to ALI, BiGAN, MDGAN, and VEEGAN}

As discussed in section~\ref{sec02}, mode collapse occurs because a generator can fool a discriminator by producing the same or similar samples corresponding to a single major mode of ${\mathit{P}}_{data}$. This issue frequently arose when ${\mathit{P}}_{data}$ includes many minor modes. Several recent studies state that traditional GANs imposing unidirectional mapping (i.e., generation mapping) is not sufficient enough to GAN training. To address this problem, they suggest bidirectional mapping to regularize generator training \cite{ref01,ref02,ref03,ref04}. Their network architecture is similar to the proposed model in that they also utilize an encoder architecture to map ${\mathit{P}}_{data}$ into low dimensional manifold space. However, while their encoders are designed to map ${\mathit{P}}_{data}$ into ${\mathit{P}}_z$ (i.e., inference mapping), we intend to map ${\mathit{P}}_{data}$ onto meaningful manifold space we determined; namely manifold mapping. To clarify the difference, we refer existing techniques as bidirectional GANs, and the proposed model as weakly bidirectional GANs, respectively.

Bidirectional mapping cannot avoid at least one of the two limitations depending on whether additional constraints are applied to the encoder or not. By applying the constraint on the encoder, the encoder loses the representational power. Meanwhile, without the constraint, the generator loses the generation power to cover the wide range of latent distribution. 
In the former case, encoder encodes ${\mathit{P}}_{data}$ by ensuring that encoded distribution matches ${\mathit{P}}_z$. This strict constraint is rarely satisfied in reality, thus increases training instability, and reduces the representational power of the encoder  \cite{ref03}. On the contrary, in the latter case, the generator is forced to handle two latent distributions ${\mathit{P}}_{z}$ and ${\mathit{P}}_{\widehat{z}}$ concurrently. The less two distributions overlap, the lower quality in generated samples from the generator. Furthermore, bidirectional mapping propagates errors from generation (or inference) mapping to inference (or generation) mapping. As a result, this becomes an additional source of training instability. 

Unlike bidirectional mapping, our weakly bidirectional mapping trains the encoder network independently from GAN training and then fixed.
Furthermore, we guarantee the generator to be focused on covering ${\mathit{P}}_z$ only because our weakly bidirectional mapping imposes manifold mapping, not inference mapping.
As a result, our weakly bidirectional mapping technique allows the encoder and the generator to be trained without strict constraints that degrade their performance.
Figure~\ref{fig02} visualizes the conceptual difference between  bidirectional mapping approach and our weakly bidirectional mapping approach. Note that both approaches try to decrease two divergences simultaneously and one of them is the divergence between ${\mathit{P}}_{data}$ and ${\mathit{P}}_{model}$. Yet, bidirectional mapping additionally considers the divergence between ${\mathit{P}}_{z}$ of the real data and ${\mathit{P}}_{z}$ of the fake data, while our mapping aggregate the divergence between ${\mathit{P}}_{m}$ of the real data and ${\mathit{P}}_{\widehat{m}}$ of the fake data.

More specifically, previous studies using bidirectional mapping employ the discriminator for joint distribution matching \cite{ref01,ref02}, reconstruction loss (i.e., pixel-wise L1 or L2 loss) \cite{ref03}, or both \cite{ref04}. 
The discriminator for joint distribution matching is to evaluate both generation and inference mapping by distinguishing between two joint distributions: the joint distribution of the real data and its inferred latent from an encoder, and that of the real latent vector and its generated data from a generator. In other words, the single discriminator should achieve two different goals;  $D$ evaluates 1) whether generated data is real or not, and 2) whether both joint distributions matches or not. Thus, their discriminator becomes insensitive to subtle changes in each distribution. Consequently, this is likely to increase mode missing that are relative minor in the data distribution \cite{ref03,ref04}. Furthermore, they do not provide any regularization or constraint condition to prevent error propagation when iterating generation and inference mapping. For these reasons, the sample reconstruction from ALI and BiGAN tend to produce less faithful reproductions of the inputs.

MDGAN and VEEGAN introduce a reconstruction loss into the GAN framework in order to prevent mode missing in ${\mathit{P}}_{data}$. 
This loss enforces each data sample $x$ (or $z$) to be reconstructed correctly after applying inference and generation mapping sequentially.
With this reconstruction loss, MDGAN and VEEGAN improve inference mapping compared to ALI and BiGAN. However, it is hard to tune parameters for balancing between adversarial loss and reconstruction loss because their units are different. (e.g., adversarial loss measures the divergence and reconstruction loss measures the pixel difference)   

Inspired by this observation, we propose an weakly bidirectional mapping approach to subsiding drawbacks from bidirectional mapping. It is important to note that our network model does not include a direct link between an encoder and a generator, meaning that training procedure of the encoder and that of the generator are separated. This separation is effective in improving the performance of the encoder and the generator respectively. This is because they can focus on their own objective without distraction by constraints. Also, because the guidance network assesses the divergence of two distributions analogous to the standard GAN loss, there is no unit mismatch to integrate the two losses.

\subsection{Characteristics of guidance network}

As shown in Fig.~\ref{fig01}, the guidance network consists of an encoder and a discriminator. In order to solve the mode collapse, we design the encoder $E$ such that the output distribution of encoder ${\mathit{P}}_m$ is the best approximate of ${\mathit{P}}_{data}$ given the fixed dimensionality of ${\mathit{P}}_m$; all modes of ${\mathit{P}}_{data}$ are reflected in ${\mathit{P}}_m$. To meet this criteria, we employ the encoder of pre-trained autoencoder.
The autoencoder first learns a representaion of a dataset using an encoder, and then reconstructs the dataset by decoding them from the representation.
Because the autoencoder network is trained to minimize the reconstruction errors (i.e., L1 or L2 loss between the input and its reconstruction), the autoencoder could learn all modes of true data distribution \cite{ref34}. It is because every sample in the dataset equally contributes to train the network, memorizing all data. Although this property causes the quality degradation of image generation (e.g., image blurs), this is advantageous to achieve the goal of the guidance network, which induce the generator to learn true distribution without missing modes.
Owing to its representational power of autoencoder, the encoder is effective to represent ${\mathit{P}}_{data}$ such that ${\mathit{P}}_m$ can reflect all modes of ${\mathit{P}}_{data}$.
Specially, we pre-train and fix the parameters of autoencoder using a real dataset. In this way, it is possible to keep the representational power of the encoder, and reduce the uncertainty of the inference. 

Since the manifold space is a topological space, a general distance measure is not suitable for the dissimilarity measure between two samples, one from ${\mathit{P}}_m$ and the other from ${\mathit{P}}_{\widehat{m}}$ \cite{ref06}. To measure the dissimilarity between ${\mathit{P}}_m$ and ${\mathit{P}}_{\widehat{m}}$, the discriminator of guidance network $D_m$ learns to separate two distributions in the manifold space based on adversarial learning. For constructing $D_m$, we use the identical structure and divergence to the discriminator of the standard GAN. Although we simply add two losses without further investigation, we achieve stable training because we are free from unit mismatch. 


\section{Evaluation}

For quantitative and qualitative evaluations, we utilize simulated and two real datasets: CelebA \cite{ref21} and CIFAR-10 \cite{ref23}, normalizing between -1 and 1. Note that the input dimensionality of CelebA is (64, 64, 3) and CIFAR-10 is (32, 32, 3). A denoising autoencoder \cite{ref31} is adopted for the guidance network to encourage robust feature extraction, resulting in a slight quality improvement compared to a conventional autoencoders. 

\subsection{Synthetic data}
\begin{figure}[t!]
  \centering
    \includegraphics[width=0.45\columnwidth]{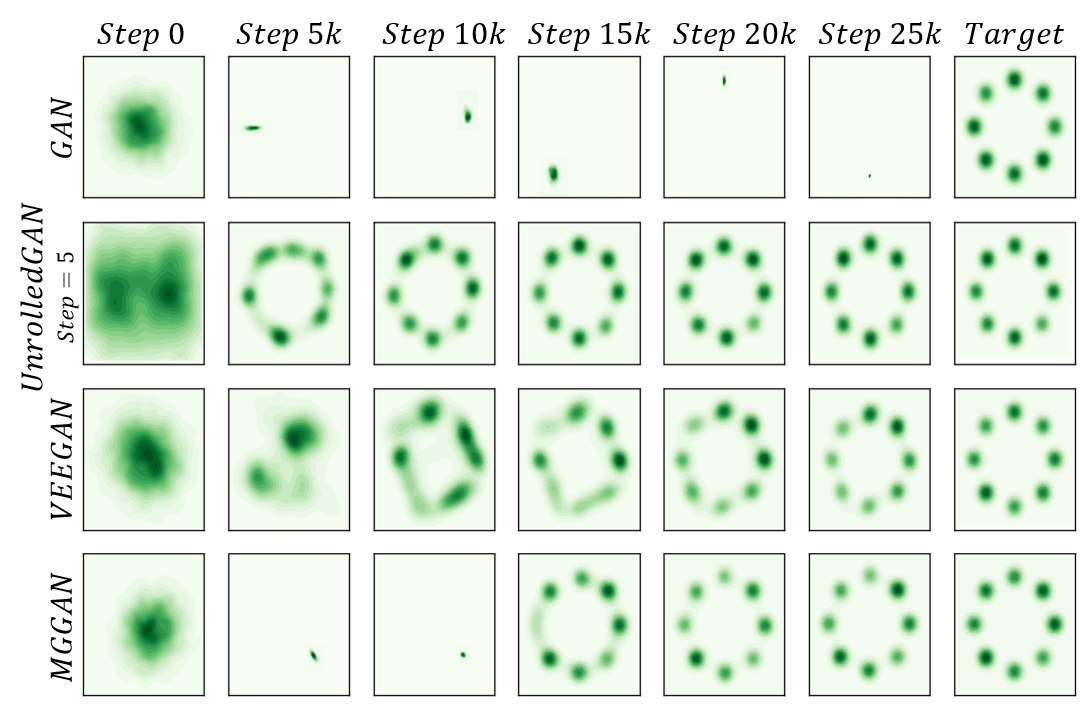}
    \qquad
    \includegraphics[width=0.45\columnwidth]{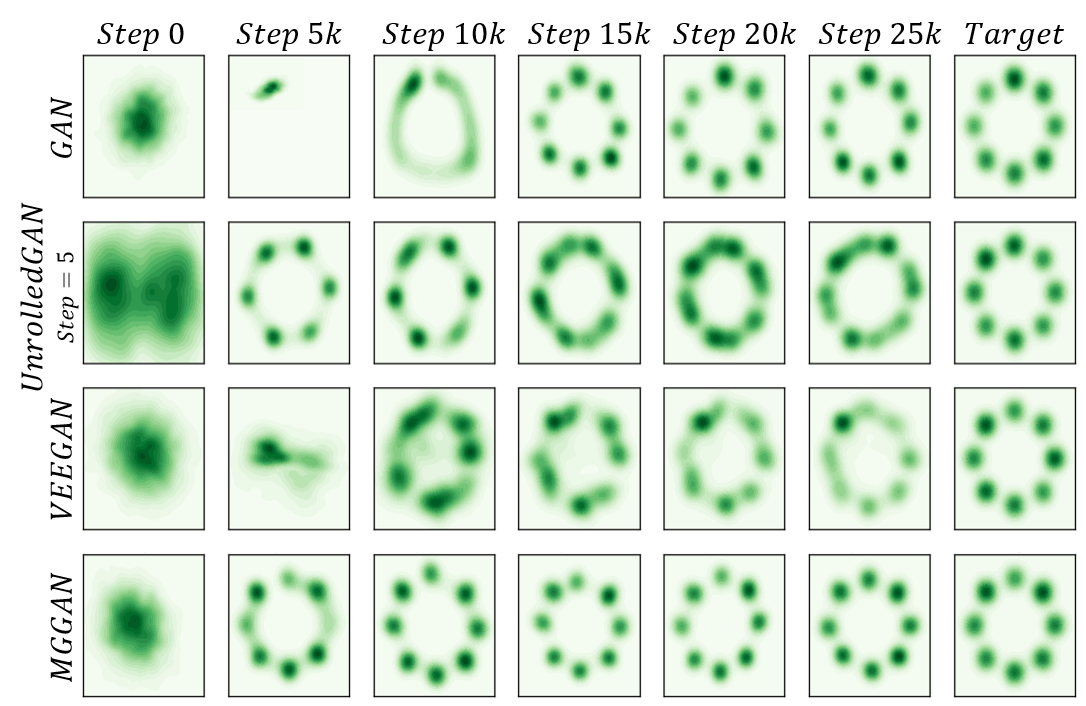}
    \caption{Mode collapse test learning a mixture of eight Gaussian spreads in a circle with 0.01 (Left) and 0.35 standard deviation (Right).}
    \label{figure03}
\end{figure}

To demonstrate that the guidance network helps GANs to prevent mode missing, i.e., solving mode collapse, we train and test the network using a simple 2D Gaussian mixture model of which eight modes are evenly distributed along the circle \cite{ref14}. We set the standard deviation (std) to 0.01 and 0.35 to see how the interval among modes influences mode collapse. Figure~\ref{figure03} compares MGGAN, GAN, unrolled GAN\footnote{ We refer the code in \textrm{https://github.com/poolio/unrolled\_gan} }, and VEEGAN\footnote{ We refer the code in \textrm{https://github.com/akashgit/VEEGAN} } models. When modes are far apart (i.e., std = 0.01), the GAN suffers from mode collapse while other models effectively solves this problem. In contrast, when the modes are adjacent (i.e., std = 0.35), unrolled GAN and VEEGAN capture almost all modes, but generate highly scattered samples that do not accurately represent the true distribution. Unlike the earlier example, the GAN outperforms both unrolled GAN and VEEGAN in the latter experiment. In both cases, our MGGAN consistently resolve mode collapse with an accurate representation.  

Interestingly, we observe that MGGAN first captures each mode, and then deviates from mode collapse: Fig.~\ref{figure03} supports this when a std is 0.01. This is because MGGAN is built upon the standard GAN, but the guidance network induces a generator to learn the entire mode. In this reason, MGGAN shows learning patterns similar to the GAN with a std of 0.35, and can generate samples of fine quality similar to the GAN.

\subsection{Quantitative evaluation}

For evaluating the effectiveness of our MGGAN, we construct four variants of MGGAN. That is, we select four different GANs as baseline networks, and then modify each by adding the guidance network. Those of baseline GANs report the state of the art visual quality in data generation, but are prone to mode collapse. Throughout this paper, we utilize four baseline networks as DCGAN \cite{ref27}, LSGAN \cite{ref28}, DRAGAN \cite{ref15}, and DFM \cite{ref29}, and develop the variants of MGGAN as DCGAN-MG, LSGAN-MG, DRAGAN-MG, and DFM-MG. For the fair comparison, the network architecture of both a generator and a discriminator follows that of DCGAN. Also, we utilize suggested hyperparameters from each baseline work without any fine-tune. Implementation code is available soon. 

MS-SSIM \cite{ref26} and the inception score \cite{ref25} are used as metrics for quantitative evaluation. These demonstrate that MGGAN improves the diversity of data generation while retaining the image quality of baseline GANs. The smaller MS-SSIM implies the better performance in producing diverse images. The inception score is used to assess the visual quality of GANs using the CIFAR-10 dataset, and the larger score represents the higher quality.  

To evaluate the image diversity using MS-SSIM, we only use the CelebA dataset. CIFAR-10 is excluded from this experiment, because MS-SSIM is meaningless if the dataset is already highly diverse \cite{ref08}; CIFAR-10 is composed of ten different classes. To compare four variants of MGGAN with their baseline GANs, we measure MS-SSIM for 100 samples generated from four baseline GANs with and without the guidance network. Table~\ref{table01} summarizes the average score of MS-SSIM measurements repeated ten times for each model. From this experiment, we find that four variants of our MGGAN significantly improves the image diversity (i.e., reduced MS-SSIM) compared to the baseline GANs all the time. Furthermore, the MS-SSIM values of all MGGANs are close to that of real data (i.e., 0.3727). This justifies that the proposed model is effective to handle the mode collapse. It is because the level of image diversity from the proposed model nearly approaches to its optimal limit, the image diversity of real dataset. Particularly, DCGAN-MG shows the most notable improvement over DCGAN, because DCGAN is the most prone to mode collapse. 

\begin{table}[t!]
\caption{Comparison of the image diversity using the MS-SSIM. Four baseline GANs and our MGGANs are compared. Note that the MS-SIMM of real dataset is 0.3727. NB: the lower the MS-SSIM, the higher the diversity.}
\label{table01}
   \begin{center}
     \begin{Large}
      \begin{sc}
        \resizebox{0.6\columnwidth}{!}{\begin{tabular}{lcccc}
        \hline
         & DCGAN & LSGAN & DRAGAN & DFM \\
        \hline
        Original    & 0.4695 & 0.3904 & 0.3934 & 0.3996 \\
        with MG     & 0.3872 & 0.3784 & 0.3899 & 0.3814 \\
        \hline
        \end{tabular}}
      \end{sc}
    \end{Large}
   \end{center}
\end{table}

\begin{figure}[t!]
  \centering
    \includegraphics[width=1.02\columnwidth]{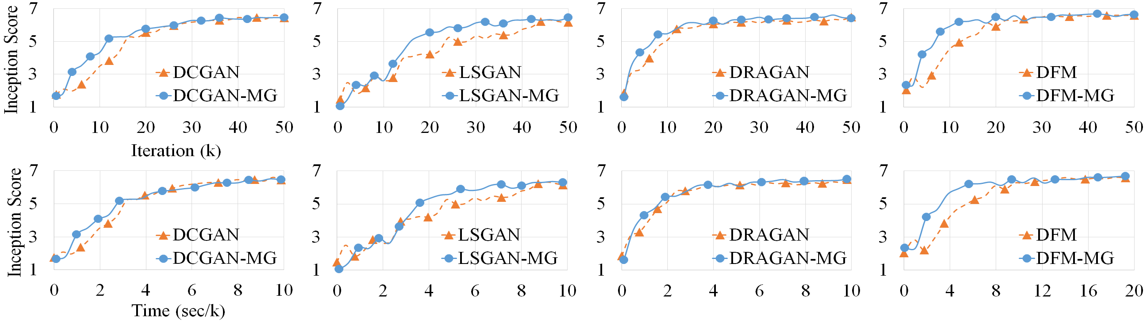}
    \vskip 0.2in
    {\Large
    \resizebox{0.6\columnwidth}{!}{\begin{tabular}{lcccc}
        \hline
         & \ DCGAN \ & \ LSGAN \ & \ DRAGAN \ & \ DFM \ \\
        \hline
        Original    & 6.4706 & 6.3243 & 6.4468 & 6.5854 \\
        with MG     & 6.4728 & 6.3416 & 6.4942 & 6.6076 \\
        \hline
        \end{tabular}}
    }
    \caption{Comparison of inception scores as a function of iteration and time. The inception scores in the table are the average scores of 5 repeated measurements of each model.}
    \label{figure04}
\end{figure}

Following Salimans et al. \cite{ref25}, we compute the inception score for 50k generated images from baseline GANs and our MGGANs. Figure~\ref{figure04} plots the inception score as a function of iteration (top) and time (bottom), respectively. We observe that the inception score from DFM is not as high as they reported in \cite{ref27}. This drop might be caused by the modification of the network architecture to DCGAN. Still, DFM marks the highest score among other GANs. From this experiment, we observe that the inception scores are not decreased in our model, and this observation holds for four different variants. More specifically, we confirm that our MGGAN can achieve the image quality of baseline GANs within approximately 0.04 tolerance of inception score. Additionally, we observe that the inception score for our MGGAN increases slightly faster than the baseline GANs. This demonstrates that a guidance network effectively accelerates the entire training by offering additional feedback of  ${\mathit{P}}_{data}$ to a generator.

\begin{figure}[t!]
  \centering
    \includegraphics[width=\columnwidth]{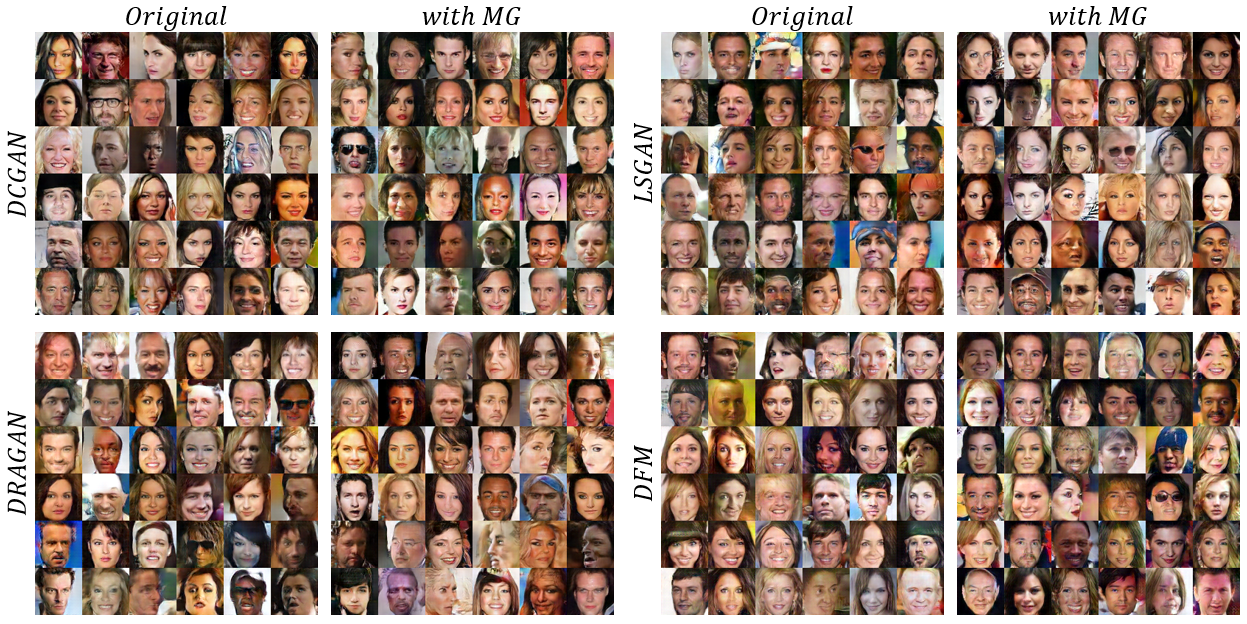}
    \caption{Comparison between randomly generated samples from original baseline GANs (DCGAN, LSGAN, DRAGAN, and DFM) and the corresponding MGGANs (DCGAN-MG, LSGAN-MG, DRAGAN-MG, and DFM-MG).}
    \label{figure05}
\end{figure}

\begin{figure}[t!]
  \centering
    \includegraphics[width=\columnwidth]{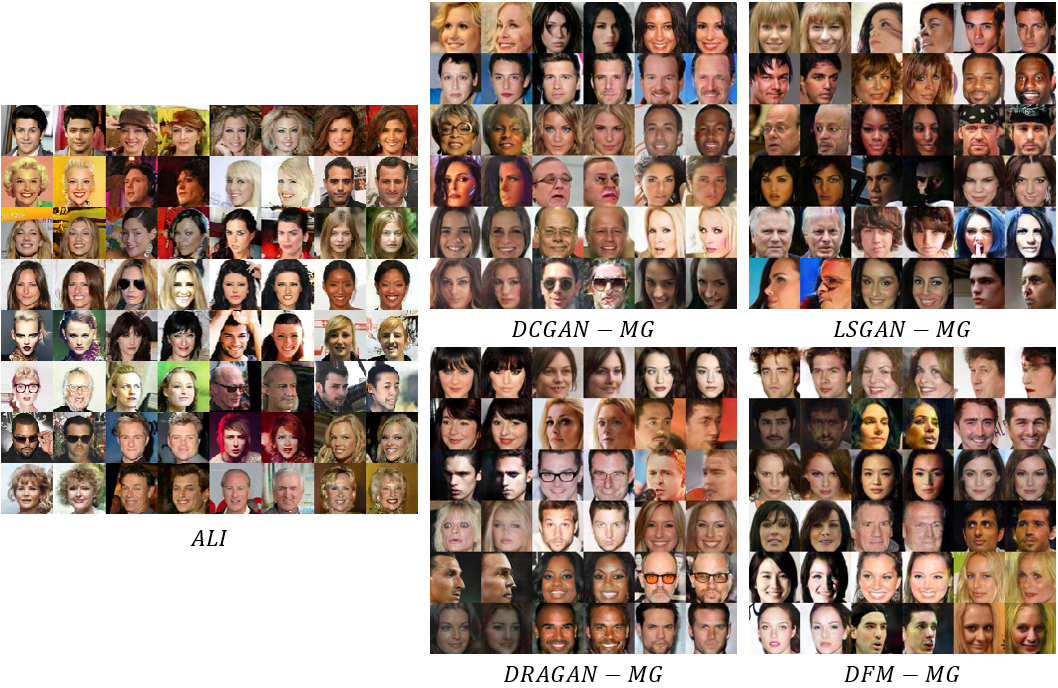}
    \caption{Reconstruction quality comparison of MGGAN variants (DCGAN-MG, LSGAN-MG, DRAGAN-MG, and DFM-MG) with ALI~\cite{ref03}. ALI results are from the paper. Odd columns are test images of CelebA dataset and even columns are corresponding reconstructions from each models.}
    \label{figure06}
\end{figure}

\subsection{Qualitative evaluation}

In this section, we investigate the effect of the guidance network whether it 1) yields the degradation in visual quality, 2) induces a meaningful manifold mapping, and 3) results in the memorization of ${\mathit{P}}_{data}$. 

First, we compare generated images from baseline models and the corresponding MGGANs. Figure~\ref{figure05} visualizes those results; the left side shows the generated images from the baseline GAN while the right side presents those from the MGGAN. From this qualitative comparison, it is hard to recognize the quality difference from both results. Therefore, our achievement in improving the image diversity is not the result of sacrificing the visual quality. These results are analogous to the quantitative evaluation reported in Fig.~\ref{figure04}.

\begin{figure}[t!]
  \centering
    \includegraphics[width=\columnwidth]{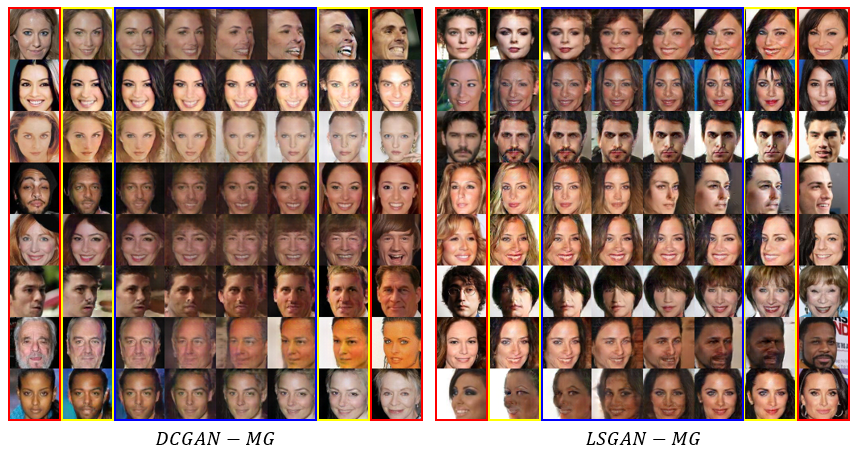}
    \includegraphics[width=\columnwidth]{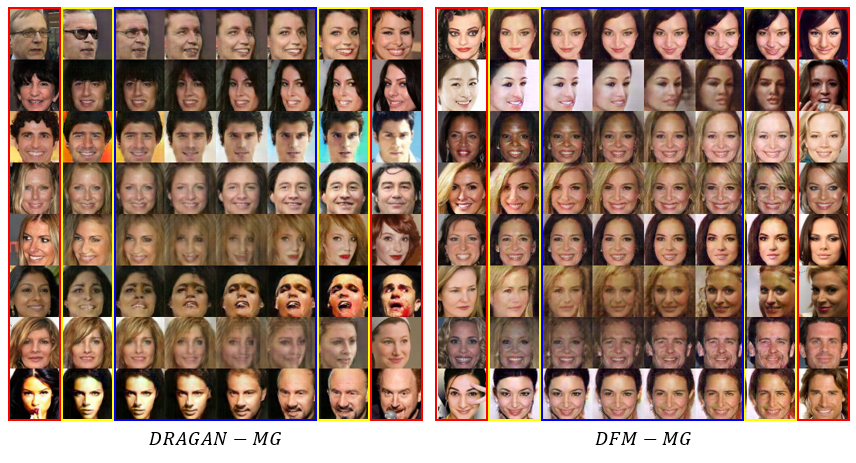}
    \caption{Latent space interpolations from CelebA dataset. Left and right-most columns, marked red box, are test images, and just besides of them, marked yellow box, are corresponding reconstructions. Intermediate columns among them, marked blue box, are linear interpolations in the latent space between reconstructions.}
    \label{figure07}
\end{figure}

Second, we exam whether our weakly bidirectional mapping can induce a meaningful cycle between data and a latent vector $z$ or not. For that, we build an additional network that associates our manifold space ${\mathit{P}}_m$ to a latent space ${\mathit{P}}_z$. Because this network transforms the encoder output to a latent vector, we could track a cyclic mapping. That is, $z$ $\Rightarrow$ $x$ $\Rightarrow$ $m$ $\Rightarrow$ $z$. This path can be considered as the detours to build a bidirectional mapping. Although this additional network is never utilized during our training, we intentionally develop this network to derive $\widehat{z}$ corresponding to $x$, and then reconstruct $x$ using the generator $G(\widehat{z})$.Based on this reconstruction experiment, we can evaluate how accurate our model can reproduce the real data, even without explicitly imposing the reconstruction loss. A network to link ${\mathit{P}}_m$ and ${\mathit{P}}_z$ is composed of 1024 full connected layer(FC) $-$ batch normalization(BN) $-$ rectified linear unit(ReLU) $-$ 1024 FC $-$ BN $-$ ReLU $-$ dimension of $P_z$ FC.
Figure~\ref{figure06} shows the reconstructed images with their target images. They are from CelebA test dataset and four variants (DCGAN-MG, LSGAN-MG, DRAGAN-MG, and DFM-MG) are all investigated. For the performance comparison with bidirectional mapping approaches, we borrow the result image of ALI from their paper \cite{ref01}. Odd columns show target images and even columns are their reconstructed images. The results from ALI do not faithfully restore the attribute of target faces, such as gender, glasses, and background color. On the contrary, our MGGANs reproduce target images reasonably well, maintaining  the original attribute. From this experiment, it is possible to confirm that our MGGAN produces more accurate reconstruction results than the bidirectional mapping approach, ALI. 

Third, we generate samples by walking in latent space to verify whether data generation is the results of data memorization or not. Because our generator learns representative features in manifold, ${\mathit{P}}_m$, derived from ${\mathit{P}}_{data}$ solely, it might be reasonable to suspect overfitting of training data. To clarify this issue, image generation results by latent walking are shown in Fig.~\ref{figure07}. Note that we choose two latent vectors, which are derived from CelebA test data using the above network (connecting the manifold to the latent space). According to Radford et al. \cite{ref27} , Bengio et al. \cite{ref32}, and Dinh et al.\cite{ref33}, the interpolated images between two images in latent space do not have meaningful connectivity when the networks just memorize the dataset: such as lack of smooth transitions or fail to generation. However, because our MGGAN produces natural interpolations with various examples, we conclude that MGGAN learn the meaningful landscape in latent space. Thus, we confirm that MGGAN does not overfit the training data.


\section{Conclusions}
In this study, we propose a new algorithm that induces a generator to produce diverse samples without sacrificing visual quality by manifold matching using the guidance network. 
To solve mode collapse, it is important to develop the manifold space where all modes of true distribution are reflected. To this end, we adopt an encoder of pre-trained autoencoder as a manifold mapping function. Because autoencoder aims to reconstruct all samples in dataset, the encoder should not ignore minor modes. Consequently, the generator avoids mode missing during training because it receives the feedback for minor modes of data distribution from the guidance network. 

Compared with existing studies of constructing bidirectional mapping in GANs training, our algorithm can be interpreted as exploiting weakly bidirectional mapping between the data and latent. Because bidirectional mapping introduces excessive constraints for network training, they lose either the generation power of a generator or representation power of a encoder. Meanwhile, the proposed algorithm utilizes manifold mapping that does not reduce the generation power of a generator but rather encourage the generation process to increase image diversity.     
Moreover, our algorithm is easily extendable to various different existing GANs. From the qualitative and quantitative experiments, we justify that MGGAN can successfully generate diverse samples without losing image quality.

In this paper, we suggest the encoder network of pre-trained autoencoder for manifold mapping in order to solve mode collapse. We believe that this idea of manifold mapping can be further extended toward integrating prior information to generator training. We hope that the weakly bidirectional mapping approach provides a basis for future work for controlling generator with prior knowledge.

\clearpage

\bibliographystyle{splncs}

\end{document}